%% file: paper.tex
\documentclass[]{TEAI}

\usepackage{algorithm}
\usepackage{algpseudocode}
\usepackage{listings}
\usepackage{xspace}
\usepackage{wrapfig}
\usepackage{caption}
\usepackage{algorithm}      
\usepackage{algpseudocode} 

\newcommand{\method}{\textsc{ARM}\xspace}
\newcommand{\bench}{\mathcal{B}}
\newcommand{\calib}{\mathcal{D}_{\mathrm{cal}}}
\newcommand{\devset}{\mathcal{D}_{\mathrm{dev}}}

\newcommand{\Tasks}{\#\,Tasks}
\newcommand{\Trajectories}{\#\,Trajectories}

\colorlet{highlight}{cyan!10}
\definecolor{gaincolor}{RGB}{0,128,0}
\definecolor{losscolor}{RGB}{180,60,60}
\definecolor{bestcolor}{RGB}{255,245,220}

\lstdefinestyle{case}{
  basicstyle=\ttfamily\scriptsize,
  columns=fullflexible,
  keepspaces=true,
  showstringspaces=false,
  breaklines=true,
  breakatwhitespace=false,
  upquote=true,
  aboveskip=2pt,
  belowskip=2pt,
  xleftmargin=0.6em
}
\newtcolorbox{casebox}[2][]{%
  enhanced,
  breakable,
  width=\columnwidth,
  colback=gray!4,
  colframe=black!25,
  colbacktitle=black!80,
  coltitle=white,
  fonttitle=\bfseries\small,
  title=#2,
  boxrule=0.45pt,
  arc=2mm,
  left=1.2mm,right=1.2mm,top=0.9mm,bottom=0.9mm,
  fontupper=\footnotesize,
  fontlower=\footnotesize,
  pad at break*=1mm,
  before skip=8pt,
  after skip=6pt,
  segmentation style={solid,black!20},
  #1
}

\title{ARM: Role-Conditioned Neuron Transplantation for Training-Free Generalist LLM Agent Merging}

\author{
Zhuoka Feng\textsuperscript{1*},
Kang Chen\textsuperscript{1*},
Sihan Zhao\textsuperscript{1**},
Kai Xiong\textsuperscript{**},
Yaoning Wang\textsuperscript{1**}\\
Minshen Yu\textsuperscript{1},
Junjie Nian\textsuperscript{1},
Changyi Xiao\textsuperscript{1},
Yixin Cao\textsuperscript{1,2$\dagger$},
Yugang Jiang\textsuperscript{1}
}

\affiliation[1]{\mbox{Fudan University}}
\affiliation[2]{\mbox{Shanghai Innovation Institute}}

\correspondence{\email{23307130211@m.fudan.edu.cn}\quad \email{yxcao@fudan.edu.cn}}
\checkdata[Website]{\url{https://arkazhuo.github.io/ARM-homepage/}}

\abstract{
\input{section/abstract}

}
\begin{document}
\maketitle

% Footnotes under title (from the ACL version)
\begingroup
\renewcommand{\thefootnote}{}
\footnotetext{* Contributed equally (Co-first authorship).}
\footnotetext{** Contributed equally (Co-second authorship).}
\footnotetext{$\dagger$ Corresponding author.}
\endgroup

\input{section/introduction}
\input{section/relatedwork}
\input{section/method}
\input{section/experiments}
\input{section/conclusion}
\input{section/discussion}

\clearpage
\ULforem  % 参考文献使用下划线
\bibliographystyle{plainnat}
\bibliography{main}
\normalem  % 恢复斜体（给 appendix 用）

\appendix
\input{section/appendix}

\end{document}

%% file: section/abstract.tex
Interactive large language model agents have advanced rapidly, but most remain specialized to a single environment and fail to adapt robustly to other environments.
Model merging offers a training-free alternative by integrating multiple experts into a single model.
In this paper, we propose Agent-Role Merging (\method), an activation-guided, role-conditioned neuron transplantation method for model merging in LLM agents.
ARM improves existing merging methods from static natural language tasks to multi-turn agent scenarios, and over the generalization ability across various interactive environments.
This is achieved with a well designed 3-step framework: 1) constructing merged backbones, 2) selection based on its role-conditioned activation analysis, and 3) neuron transplantation for fine-grained refinements.
Without gradient-based optimization, ARM improves cross-benchmark generalization while enjoying efficiency.
{Across diverse domains, the model obtained via ARM merging outperforms prior model merging methods and domain-specific expert models, while demonstrating strong out-of-domain generalization.}

%% file: section/introduction.tex
\section{Introduction}
\label{sec:intro}

Recently, we have witnessed the surge of agents by fine-tuning large language models (LLMs) in interactive environments, such as web browsing and operating systems~\citep{liu2023agentbench,zheng2025lifelongagentbench}.
These LLM-based agents can think, plan, and act through external tools to accomplish real-world tasks, making them practically valuable~\citep{yao2023react,qin2024toolllm}. 

Despite these advances, current LLM-based agents often exhibit limited cross-environment robustness~\citep{yao2024taubench,wang2024officebench}.
Models tuned for one environment often degrade sharply when deployed in another one with different tool schemas, action interfaces, or trajectory distributions~\citep{yao2024taubench,wang2024officebench}.
A straightforward solution is to further fine-tune a single model across all  environments, but this introduces substantial engineering and optimization complexity (e.g., curriculum/order effects across environments, heterogeneous tool interfaces, and extensive debugging) and incurs huge training costs~\citep{agentrl2025,chainofagents2025}.

In this paper, we focus on a training-free alternative, model merging. It aims at combining multiple checkpoints of the same architecture --- often specialists fine-tuned for different environments (marked as expert in the rest of paper) --- into a single model that aims to inherit the strengths of each~\citep{ilharco2023task_arithmetic,yadav2023ties}. Without additional training, model merging offers a practical path to greatly improving capabilities and reducing the burden of maintaining many specialized checkpoints~\citep{wortsman2022model_soups,ilharco2023task_arithmetic}. 
A growing literature studies training-free merging, from simple parameter-space compositions \citep{ilharco2023task_arithmetic} to interference-aware recipes such as TIES-Merging \citep{yadav2023ties}. Recent work further investigates activation merging (e.g., AIM \citep{aim2025}, NeuronMerge \citep{neuronmerge2025}), leveraging internal signal tracing to mitigate inter-model interference.
However, the above methods are predominantly developed on static, single-turn tasks, and few works target interactive agent settings.

\begin{wrapfigure}[16]{r}{0.48\columnwidth}
  \centering
  \includegraphics[width=\linewidth]{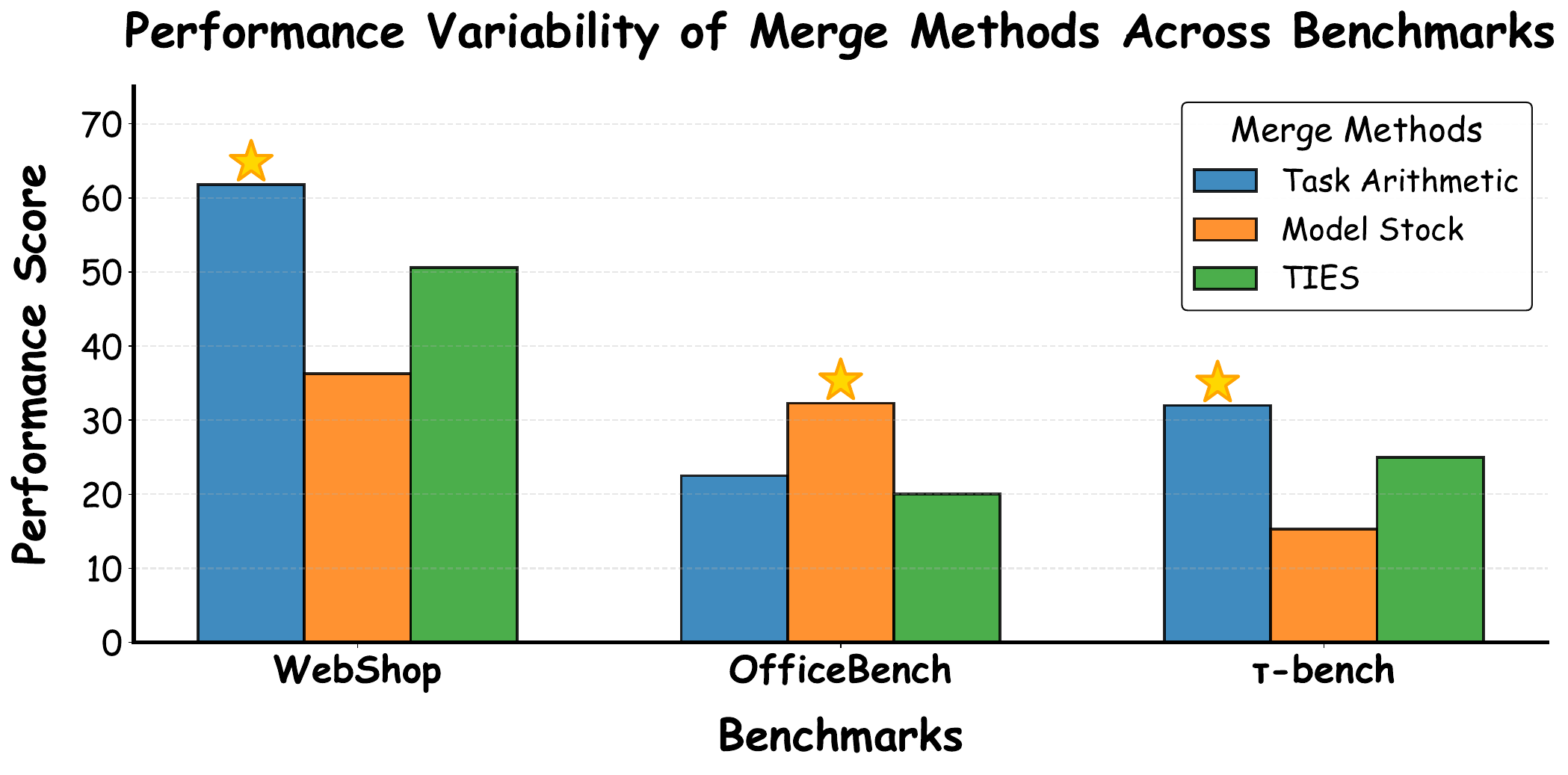}
  \caption{Performance variability of common training-free merge heuristics across interactive agent benchmarks.}
  \label{fig:merge_robustness}
\end{wrapfigure}
To this end, we propose agentic merging to combine multiple LLMs that generalize well across interactive environments. We highlight two critical challenges. First, how can we preserve general capabilities reliably? 
Different base model families exhibit different internal mechanism and activation features. 
Their behaviors can become highly unstable across benchmarks.
As shown in Figure~\ref{fig:merge_robustness}, widely used heuristics exhibit pronounced cross-benchmark variance, with no single heuristic consistently strong across environments.
This motivates a stability-aware backbone selection strategy prior to any fine-grained intervention.
Second, how can we avoid capability conflicts during merging? This is a core challenge for model merging, and multi-turn agent trajectories exacerbate it. Small deviations in role-critical spans (e.g., tool-call formatting, action serialization, or final-answer JSON) can cascade into repeated failures and negative transfer across environments.

Therefore, we design Agent-Role Merging (ARM), an activation-guided, role-conditioned neuron transplantation framework for training-free model merging. To tackle the first challenge, ARM introduces a dynamic backbone construction and selection stage. It constructs a small candidate pool of merged backbones using standard weight-space merge operators, then dynamically selects a strong one using a well-designed strategy based on mechanism analysis. Note that this step remains training-free and avoids costly operations while maximizing the reserved capabilities. 
To tackle the second challenge, ARM performs fine-grained neuron transplantation at the level of role-critical behaviors. Specifically, ARM conducts role-conditioned activation tracing on a small calibration set to identify key neurons for specific abilities (e.g., tool calls, actions, and final-answer JSON), and then selectively transplants these neurons from the corresponding expert into the chosen backbone. We also use a conflict-aware policy to reduce negative transfer in multi-turn settings.

We evaluate ARM on multiple widely used agentic benchmarks, and show that it yields the strongest single merged generalist across both Qwen3 and Qwen2.5 expert pools, improving average performance and worst-suite robustness while maintaining strong out-of-domain generalization compared to prior training-free merging baselines.

Our contributions are summarized as follows:
\begin{itemize}
  \item We propose to curate and select merged backbones dynamically for reliable general capability reservation.
  \item We propose a fine-grained neuron transplantation mechanism for agentic LLM merging towards better generalization.
  \item Extensive experiments on four in-domain suites and two out-of-domain benchmarks demonstrate that ARM consistently improves generalist performance and robustness over strong weight-space and activation-aware training-free baselines using a single merged checkpoint.
\end{itemize}

%% file: section/relatedwork.tex
\section{Related Work}
\label{sec:related}

\paragraph{Training generalist agents.}
One route to cross-environment generalization is to train a single generalist agent using large-scale multi-task trajectories, often via online interaction and reinforcement learning.
AgentRL \citep{agentrl2025} explores scaling agentic RL in multi-turn settings, and Chain-of-Agents \citep{chainofagents2025} studies multi-agent distillation and agentic RL for foundation agents.
While effective, such pipelines require expensive interaction and task coverage; our goal is complementary: training-free composition of existing specialists.

\paragraph{Model merging beyond static-task heuristics.}
Model merging combines multiple fine-tuned models into a single model without additional gradient updates, ranging from weight averaging and model soups \citep{wortsman2022model_soups}, task arithmetic and task vectors \citep{ilharco2023task_arithmetic} to interference-aware recipes such as TIES-Merging \citep{yadav2023ties}.
Recent work also explores activation-aware merging, such as AIM \citep{aim2025} and NeuronMerge \citep{neuronmerge2025}, to mitigate interference by tracing internal signals.
However, most existing approaches are developed and validated on static, single-turn NLP tasks. They lack activation-based criteria for selecting a strong merged backbone and do not incorporate conflict-aware policies to protect role-critical circuits when importing benchmark-specific behaviors.
Our framework complements these efforts by using role-conditioned activation tracing for backbone selection and conflict-aware, role-salient neuron transplantation for mitigating negative transfer in interactive agents.

%% file: section/method.tex
\section{Method}
\label{sec:method}

In this section, we present Agent-Role Merging~(\method), a training-free pipeline for consolidating benchmark-specialized experts into a single multi-benchmark agent model.
\method proceeds in three phases: (i)~\textbf{Backbone Pool Construction}, which constructs a pool of merged backbones via training-free weight-space merging, (ii)~\textbf{Backbone Selection} selects the backbone that best preserves expert role-salient neurons using Activation-Overlap Score (AOS), and (iii)~\textbf{Neuron Transplantation} repairs remaining capability gaps via conflict-aware neuron transplantation while strictly protecting neurons that are important for any other benchmark.

\begin{figure*}[t]
  \centering
  \includegraphics[width=\textwidth]{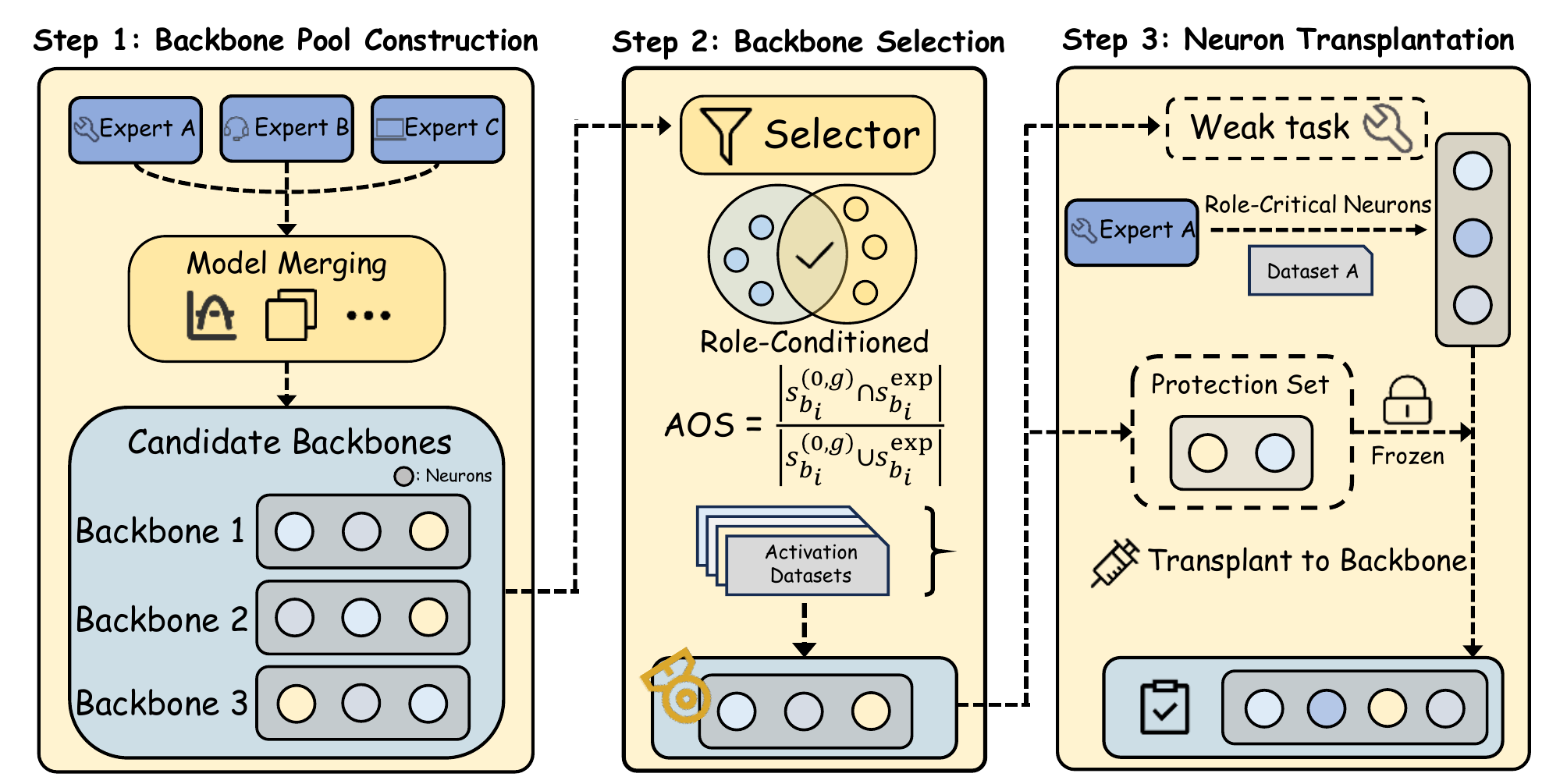}
  \caption{\textbf{Overview of Agent-Role Merging (ARM).}
  \textbf{Step 1: Backbone pool construction.} We apply multiple training-free weight-space merge operators to benchmark-specialized experts to obtain a pool of candidate merged backbones.
  \textbf{Step 2: Backbone selection.} A selector computes the \emph{Activation-Overlap Score (AOS)} using role-conditioned MLP activations on a lightweight calibration set, and chooses the candidate backbone that maximizes mean AOS across benchmarks.
  \textbf{Step 3: Neuron transplantation.} For benchmarks where the selected backbone remains weak, we transplant a small top-$k\%$ subset of donor (expert) MLP neurons into the backbone while strictly protecting neurons salient for other benchmarks to avoid negative transfer.
  The resulting single model consolidates expert capabilities across benchmarks without end-to-end retraining.}
  \label{fig:figuremain}
\end{figure*}

\subsection{Backbone Pool Construction}
\label{sec:backbone_pool}

To compare different merging strategies and select the best backbone, we first need to construct a set of candidate backbones, where each backbone represents a single merged model. Thus, we consider $N$ benchmark-specialized experts $\{M^{\mathrm{exp}}_{b_i}\}_{i=1}^N$ fine-tuned from the same base LLM, and thus sharing the same architecture and tokenizer. $\bench$ denotes the set of benchmarks, and $M^{\mathrm{exp}}_{b_i}$ denote the expert corresponding to the $i$-th benchmark $b_i \in \bench$ with a one-to-one mapping between benchmarks and experts. Our goal is to obtain a single merged model that performs well across all benchmarks in $\bench$ without additional gradient-based retraining. Since different training-free weight-space merge operators can yield markedly different trade-offs, we construct a small pool of candidate merged backbones by applying a set of standard merging operators $\mathcal{G}$, such as uniform averaging~\citep{wortsman2022model_soups}, task arithmetic~\citep{ilharco2023task_arithmetic}, and TIES-Merging~\citep{yadav2023ties}. Each operator $g \in \mathcal{G}$ produces a candidate backbone:
\begin{equation}
M^{(0,g)} \;=\; g\!\left(\{M^{\mathrm{exp}}_{b_i}\}_{i=1}^N\right).
\end{equation}

\subsection{Backbone Selection}
\label{sec:selection}

To select the backbone that best preserves expert role-salient neurons for model merging, we first need to compare candidate merged backbones, which requires a criterion that identifies the parameters supporting benchmark-critical behaviors. 

To achieve this, we propose role-conditioned MLP activations, which serve as a lightweight, training-free criterion, to approximate these circuits and summarize them as top-$k$ neuron sets.

Specifically, a small calibration set $\calib$ is sampled from splits that are disjoint from evaluation/test sets~(see Section~\ref{sec:exp}). $\calib^{(b_i)} \subseteq \calib$ denote the calibration trajectories for benchmark $b_i$.

For a given model $M$ with $L$ blocks, $\mathbf{z}_{\ell}(t) \in \mathbb{R}^{d_{\mathrm{ff}}}$ denotes the post-activation vector of the MLP in block $\ell$ at token position $t$. $T_{b_i,r}(x)$ denotes the role-$r$ token positions in trajectory $x$ for benchmark $b_i$.
Trajectories without a given role segment (i.e., $T_{b_i,r}(x)=\emptyset$) are ignored when estimating saliency for $(b_i,r)$.
Therefore, Role-conditioned saliency is defined as the expected per-role mean activation:
\begin{equation}
\small
\label{eq:saliency}
\begin{aligned}
s_{\ell,j}(M; b_i,r)
&= \mathbb{E}_{x \sim \calib^{(b_i)}}\!\left[
\operatorname*{mean}_{t \in T_{b_i,r}(x)} \bigl| z_{\ell,j}(t) \bigr|
\right].
\end{aligned}
\end{equation}

Next, the top-$k$ fraction of neurons is selected per layer:
\begin{equation}
\small
\label{eq:topk}
\begin{aligned}
S_{\ell}(M; b_i,r)
&= \textsc{TopK}_{j}\!\left(
s_{\ell,j}(M;b_i,r),\;
\left\lceil k\,d_{\mathrm{ff}} \right\rceil
\right),
\end{aligned}
\end{equation}
where $S(M;b_i,r)=\{(\ell,j): j \in S_{\ell}(M;b_i,r)\}$ is the role-salient set.
An element $n=(\ell,j)\in S(M;b_i,r)$ is referred to as a neuron index.
For brevity, $S(M;b_i)\equiv S(M;b_i,r_{b_i})$ is used when the target role is clear from the benchmark.

Hereafter, we design an \emph{Activation-Overlap Score (AOS)} to select a backbone that preserves role-salient neurons from the corresponding experts across benchmarks.
Specifically, for each benchmark $b_i$, we denote the role-salient neuron sets of the expert model and the candidate backbone as
$S^{\mathrm{exp}}_{b_i}$ and $S^{(0,g)}_{b_i}$, respectively.

Based on these definitions, the Activation-Overlap Score of a candidate backbone on benchmark $b_i$ is defined as:
\begin{equation}
\label{eq:aos}
\mathrm{AOS}(M^{(0,g)}; b_i)
=
\frac{
\left|
S^{(0,g)}_{b_i}
\cap
S^{\mathrm{exp}}_{b_i}
\right|
}{
\left|
S^{(0,g)}_{b_i}
\cup
S^{\mathrm{exp}}_{b_i}
\right|
}.
\end{equation}

Finally, we select the backbone with the highest mean AOS:
\begin{equation}
\label{eq:backbone_select}
\begin{aligned}
g^\star &= \arg\max_{g \in \mathcal{G}}\frac{1}{|\bench|}\sum_{b_i \in \bench} \mathrm{AOS}(M^{(0,g)}; b_i),\\
M^{(0)} &= M^{(0,g^\star)}.
\end{aligned}
\end{equation}
Using the AOS, the selector favors backbones that best preserve experts’ role-salient neurons, yielding a robust starting point without exhaustive evaluation of every merge candidate.

\subsection{Neuron Transplantation}
\label{sec:transplantation}

To repair remaining capability gaps and protect neurons that are important for any other benchmarks, we propose conflict-aware neuron transplantation.

Specifically, we first conduct capability-gap diagnosis to identify weak benchmarks $\bench_{\mathrm{weak}}\subseteq\bench$ introduced by role-specific regressions in model merging. A held-out development set $\devset$~(e.g., benchmarks with the largest performance gaps between $M^{(0)}$ and the corresponding expert) is selected to apply transplantation for $b_i\in\bench_{\mathrm{weak}}$. This development evaluation is used only to decide where transplantation is applied (not to train parameters). For each benchmark $b_i$, we use its corresponding expert as the donor and denote it by $M^{\mathrm{don}}_{b_i} \equiv M^{\mathrm{exp}}_{b_i}$.

Next, considering that weight-space merging is a global operation and can blur or overwrite benchmark-specific circuits.
To correct specific failures without retraining the entire model, we perform localized edits by transplanting a small number of donor MLP neurons into the selected backbone. To achieve this, we refine $M^{(0)}$ by selectively transplanting a small subset of MLP neurons from donors into the backbone.
For block $\ell$, the MLP parameters are denoted as
$W^{\ell}_{\mathrm{in}} \in \mathbb{R}^{d_{\mathrm{ff}}\times d}$, in which
$b^{\ell}_{\mathrm{in}} \in \mathbb{R}^{d_{\mathrm{ff}}}$,
and $W^{\ell}_{\mathrm{out}} \in \mathbb{R}^{d \times d_{\mathrm{ff}}}$.
Neuron $(\ell,j)$ corresponds to row $j$ of $W^{\ell}_{\mathrm{in}}$, entry $j$ of $b^{\ell}_{\mathrm{in}}$, and column $j$ of $W^{\ell}_{\mathrm{out}}$.
A hard transplantation from donor $M^{\mathrm{don}}$ into model $M$ performs:
\begin{equation}
\begin{aligned}
W^{\ell}_{\mathrm{in}}[j,:] &\gets W^{\ell,\mathrm{don}}_{\mathrm{in}}[j,:],\\
b^{\ell}_{\mathrm{in}}[j]   &\gets b^{\ell,\mathrm{don}}_{\mathrm{in}}[j],\\
W^{\ell}_{\mathrm{out}}[:,j] &\gets W^{\ell,\mathrm{don}}_{\mathrm{out}}[:,j],
\end{aligned}
\end{equation}
\begin{wrapfigure}{r}{0.50\columnwidth} 
  \vspace{-0.2\baselineskip}
  \begin{minipage}{\linewidth}
    \scriptsize

    % ---- 顶部“算法标题栏”（手动）----
    \hrule height 0.6pt
    \vspace{0.25ex}
    \refstepcounter{algorithm}%
    \textbf{Algorithm~\thealgorithm: \method --- Candidate merging, backbone selection, and conflict-aware neuron transplantation}%
    \label{alg:armerge}
    \vspace{0.2ex}
    \hrule height 0.4pt
    \vspace{0.35ex}

    % ---- 算法正文 ----
    \begin{algorithmic}[1]
      \Require Expert models $\{M^{(i)}\}_{i=1}^{N}$; merge operators $\mathcal{G}$; benchmarks $\bench$.
      \Ensure Merged model $M^\star$.
      \State Set donor $M^{\mathrm{don}}_b \gets M^{\mathrm{exp}}_b$ for each $b \in \bench$.
      \State Compute donor saliency $S^{\mathrm{don}}_b \gets S(M^{\mathrm{don}}_b; b)$ for each $b \in \bench$.
      \State Construct candidate backbones $\{M^{(0,g)}\}_{g \in \mathcal{G}}$.
      \For{$g \in \mathcal{G}$}
        \State Compute backbone saliency $S^{(0,g)}_b \gets S(M^{(0,g)}; b)$ for each $b \in \bench$.
        \State $\mathrm{Score}(g) \gets \frac{1}{|\bench|}\sum_{b \in \bench} \mathrm{AOS}(M^{(0,g)}; b)$.
      \EndFor
      \State $g^\star \gets \arg\max_{g \in \mathcal{G}} \mathrm{Score}(g)$.
      \State $M \gets M^{(0,g^\star)}$.
      \State Assign backbone saliency $S(M;b) \gets S^{(0,g^\star)}_b$ for each $b \in \bench$.
      \State Select weak benchmarks $\bench_{\mathrm{weak}} \subseteq \bench$ on $\devset$.
      \For{$b \in \bench_{\mathrm{weak}}$}
        \State $\mathcal{P}_{-b} \gets \bigcup_{b' \in \bench,\; b'\neq b} S(M;b')$.
        \State $\mathcal{T}_b \gets \{n \in S^{\mathrm{don}}_b \mid n \notin \mathcal{P}_{-b}\}$.
        \State Transplant neurons in $\mathcal{T}_b$ from $M^{\mathrm{don}}_b$ into $M$.
      \EndFor
      \State \textbf{return} $M$.
    \end{algorithmic}

    % ---- 底部线 ----
    \vspace{0.25ex}
    \hrule height 0.6pt
  \end{minipage}
  \vspace{-0.5\baselineskip}
\end{wrapfigure}
\noindent
For gated MLPs (e.g., SwiGLU), a neuron index $j$ corresponds to the same index across the gate and up projections as well as the down projection; we transplant the corresponding rows and columns accordingly.
 We apply transplantation only to a small set of neurons, keeping the rest of the network unchanged to reduce collateral interference, similar in spirit to localized editing methods that aim to confine behavioral changes \citep{meng2022rome, meng2023memit}.
 
Finally, since naively transplanting all donor-salient neurons can overwrite neurons that the backbone already relies on for other benchmarks, causing negative transfer, we employ conflict-aware transplantation policy to \emph{strictly} protect backbone neurons that are salient for any \emph{other} benchmark, and only transplant donor neurons that do not belong to those protected sets.

For each benchmark $b_i$, we define the set of backbone neurons that are salient for any other benchmark's critical role:
\begin{equation}
\label{eq:protected_minus}
\mathcal{P}_{-b_i} \;=\; \bigcup_{b_i' \in \bench,\; b_i'\neq b_i} S(M^{(0)}; b_i'),
\end{equation}
where we use the shorthand $S(M;b_i)\equiv S(M;b_i,r_{b_i})$.
When repairing benchmark $b$, we start from the donor's role-salient neurons $S(M^{\mathrm{don}}_b; b)$ and exclude all neurons that are salient for any other benchmark:
\begin{equation}
\label{eq:transplant_set}
\begin{aligned}
\mathcal{T}_b
&= \left\{ n \in S(M^{\mathrm{don}}_b; b)\;\middle|\;
n \notin \mathcal{P}_{-b}
\right\}.
\end{aligned}
\end{equation}
With this conflict-aware transplantation strategy, we target capability gaps while protecting the stability of other benchmarks and minimizing negative transfer.

%% file: section/experiments.tex
\section{Experiments}
\label{sec:exp}

\begin{table*}[!htbp]
\centering
\small
\setlength{\tabcolsep}{4.2pt}
\renewcommand{\arraystretch}{1.1}

\begin{tabular}{lccccccc}
\toprule
\textbf{Model} & WebShop & OS & $\tau$-bench & OfficeBench & DB & AlfWorld & \textsc{Avg} \\
\midrule
Qwen3-8B & 30.5 & 13.2 & 32.0 & 2.9 & 41.3 & 24.0 & 24.0 \\
\midrule
Simia-Tau & 44.8 & 16.0 & \cellcolor{bestcolor}43.8 & 2.9 & 45.7 & 32.0 & 30.9 \\
Simia-OfficeBench & 51.1 & 25.7 & 16.1 & \cellcolor{bestcolor}37.5 & 43.7 & 44.0 & 36.4 \\
Simia-AgentBench & \cellcolor{bestcolor}64.8 & \cellcolor{bestcolor}29.2 & 15.9 & 3.9 & 45.7 & 42.0 & 33.6 \\
\midrule
BEST-of-Three (oracle) & 64.8 & 29.2 & 43.8 & 37.5 & 45.7 & 44.0 & 44.2 \\
\midrule
Average & \textbf{63.8} {\scriptsize\textcolor{losscolor}{(-1.5\%)}} & 27.1 {\scriptsize\textcolor{losscolor}{(-7.2\%)}} & 19.1 {\scriptsize\textcolor{losscolor}{(-56.4\%)}} & \textbf{49.8} {\scriptsize\textcolor{gaincolor}{(+32.8\%)}} & 44.7 {\scriptsize\textcolor{losscolor}{(-2.2\%)}} & 46.0 {\scriptsize\textcolor{gaincolor}{(+4.5\%)}} & 41.8 {\scriptsize\textcolor{losscolor}{(-5.4\%)}} \\
Model Stock & 36.3 {\scriptsize\textcolor{losscolor}{(-44.0\%)}} & 15.3 {\scriptsize\textcolor{losscolor}{(-47.6\%)}} & 32.3 {\scriptsize\textcolor{losscolor}{(-26.3\%)}} & 3.9 {\scriptsize\textcolor{losscolor}{(-89.6\%)}} & 43.0 {\scriptsize\textcolor{losscolor}{(-5.9\%)}} & 22.0 {\scriptsize\textcolor{losscolor}{(-50.0\%)}} & 25.5 {\scriptsize\textcolor{losscolor}{(-42.3\%)}} \\
Task Arithmetic & 61.8 {\scriptsize\textcolor{losscolor}{(-4.6\%)}} & 27.8 {\scriptsize\textcolor{losscolor}{(-4.8\%)}} & 22.5 {\scriptsize\textcolor{losscolor}{(-48.6\%)}} & 15.5 {\scriptsize\textcolor{losscolor}{(-58.7\%)}} & 38.3 {\scriptsize\textcolor{losscolor}{(-16.2\%)}} & 10.0 {\scriptsize\textcolor{losscolor}{(-77.3\%)}} & 29.3 {\scriptsize\textcolor{losscolor}{(-33.7\%)}} \\
TIES & 50.6 {\scriptsize\textcolor{losscolor}{(-21.9\%)}} & 25.0 {\scriptsize\textcolor{losscolor}{(-14.4\%)}} & 20.0 {\scriptsize\textcolor{losscolor}{(-54.3\%)}} & 43.3 {\scriptsize\textcolor{gaincolor}{(+15.5\%)}} & 46.3 {\scriptsize\textcolor{gaincolor}{(+1.3\%)}} & 32.0 {\scriptsize\textcolor{losscolor}{(-27.3\%)}} & 36.2 {\scriptsize\textcolor{losscolor}{(-18.1\%)}} \\
TIES+DARE & 54.7 {\scriptsize\textcolor{losscolor}{(-15.6\%)}} & 20.8 {\scriptsize\textcolor{losscolor}{(-28.8\%)}} & 23.6 {\scriptsize\textcolor{losscolor}{(-46.1\%)}} & 18.3 {\scriptsize\textcolor{losscolor}{(-51.2\%)}} & 37.0 {\scriptsize\textcolor{losscolor}{(-19.0\%)}} & 14.0 {\scriptsize\textcolor{losscolor}{(-68.2\%)}} & 28.1 {\scriptsize\textcolor{losscolor}{(-36.4\%)}} \\
WIDEN & 59.0 {\scriptsize\textcolor{losscolor}{(-9.0\%)}} & 29.0 {\scriptsize\textcolor{losscolor}{(-0.7\%)}} & 22.4 {\scriptsize\textcolor{losscolor}{(-48.9\%)}} & 22.5 {\scriptsize\textcolor{losscolor}{(-40.0\%)}} & 36.3 {\scriptsize\textcolor{losscolor}{(-20.6\%)}} & 14.0 {\scriptsize\textcolor{losscolor}{(-68.2\%)}} & 30.5 {\scriptsize\textcolor{losscolor}{(-31.0\%)}} \\
\midrule
NeuronMerge & 32.7 {\scriptsize\textcolor{losscolor}{(-49.5\%)}} & 16.2 {\scriptsize\textcolor{losscolor}{(-44.5\%)}} & \textbf{32.7} {\scriptsize\textcolor{losscolor}{(-25.3\%)}} & 2.9 {\scriptsize\textcolor{losscolor}{(-92.3\%)}} & 40.3 {\scriptsize\textcolor{losscolor}{(-11.8\%)}} & 28.0 {\scriptsize\textcolor{losscolor}{(-36.4\%)}} & 25.5 {\scriptsize\textcolor{losscolor}{(-42.3\%)}} \\
AIM & 61.7 {\scriptsize\textcolor{losscolor}{(-4.8\%)}} & 31.2 {\scriptsize\textcolor{gaincolor}{(+6.8\%)}} & 22.6 {\scriptsize\textcolor{losscolor}{(-48.4\%)}} & 38.0 {\scriptsize\textcolor{gaincolor}{(+1.3\%)}} & \textbf{47.7} {\scriptsize\textcolor{gaincolor}{(+4.4\%)}} & 36.0 {\scriptsize\textcolor{losscolor}{(-18.2\%)}} & 39.5 {\scriptsize\textcolor{losscolor}{(-10.6\%)}} \\
\rowcolor{highlight}\method(ours) & 62.9 {\scriptsize\textcolor{losscolor}{(-2.9\%)}} & \textbf{35.4} {\scriptsize\textcolor{gaincolor}{(+21.2\%)}} & 28.5 {\scriptsize\textcolor{losscolor}{(-34.9\%)}} & 44.9 {\scriptsize\textcolor{gaincolor}{(+19.7\%)}} & \textbf{47.7} {\scriptsize\textcolor{gaincolor}{(+4.4\%)}} & \textbf{48.0} {\scriptsize\textcolor{gaincolor}{(+9.1\%)}} & \textbf{44.6} {\scriptsize\textcolor{gaincolor}{(+0.9\%)}} \\
\bottomrule
\end{tabular}

\caption{Main results with Qwen3-8B experts. $\tau$-bench and OfficeBench report suite averages. WebShop and OS are AgentBench tasks. DB-bench and AlfWorld are \textbf{out-of-domain} benchmarks. \textsc{Avg} is the mean over the six aggregates (BEST-of-Three is an oracle expert selector baseline). Parentheses show relative change compared to BEST-of-Three for each aggregate. Best merged model per column is bold; \method is highlighted.}
\label{tab:qwen3_results}

\vspace{0.5cm}

\begin{tabular}{lccccccc}
\toprule
\textbf{Model} & WebShop & OS & $\tau$-bench & OfficeBench & DB & AlfWorld & \textsc{Avg} \\
\midrule
Qwen2.5-7B-Instruct & 55.9 & 31.2 & 18.1 & 24.2 & 50.3 & 58.0 & 39.6 \\
\midrule
Simia-Tau & 44.3 & 12.5 & \cellcolor{bestcolor}28.7 & 2.9 & 37.7 & 18.0 & 24.0 \\
Simia-OfficeBench & 45.1 & 18.8 & 12.6 & \cellcolor{bestcolor}40.0 & 36.3 & 44.0 & 32.8 \\
Simia-AgentBench & \cellcolor{bestcolor}65.6 & \cellcolor{bestcolor}31.2 & 21.4 & 2.9 & 23.0 & 10.0 & 25.7 \\
\midrule
BEST-of-Three (oracle) & 65.6 & 31.2 & 28.7 & 40.0 & 37.7 & 44.0 & 41.2 \\
\midrule
Average & \textbf{67.2} {\scriptsize\textcolor{gaincolor}{(+2.4\%)}} & 30.8 {\scriptsize\textcolor{losscolor}{(-1.3\%)}} & 18.9 {\scriptsize\textcolor{losscolor}{(-34.1\%)}} & 4.8 {\scriptsize\textcolor{losscolor}{(-88.0\%)}} & 38.0 {\scriptsize\textcolor{gaincolor}{(+0.8\%)}} & 48.0 {\scriptsize\textcolor{gaincolor}{(+9.1\%)}} & 34.6 {\scriptsize\textcolor{losscolor}{(-16.0\%)}} \\
Model Stock & 63.0 {\scriptsize\textcolor{losscolor}{(-4.0\%)}} & 31.9 {\scriptsize\textcolor{gaincolor}{(+2.2\%)}} & 16.4 {\scriptsize\textcolor{losscolor}{(-42.9\%)}} & 36.4 {\scriptsize\textcolor{losscolor}{(-9.0\%)}} & 51.0 {\scriptsize\textcolor{gaincolor}{(+35.3\%)}} & 62.0 {\scriptsize\textcolor{gaincolor}{(+40.9\%)}} & 43.5 {\scriptsize\textcolor{gaincolor}{(+5.6\%)}} \\
Task Arithmetic & 66.3 {\scriptsize\textcolor{gaincolor}{(+1.1\%)}} & 29.9 {\scriptsize\textcolor{losscolor}{(-4.2\%)}} & 14.5 {\scriptsize\textcolor{losscolor}{(-49.5\%)}} & 6.8 {\scriptsize\textcolor{losscolor}{(-83.0\%)}} & 41.7 {\scriptsize\textcolor{gaincolor}{(+10.6\%)}} & 42.0 {\scriptsize\textcolor{losscolor}{(-4.5\%)}} & 33.5 {\scriptsize\textcolor{losscolor}{(-18.7\%)}} \\

TIES & 62.0 {\scriptsize\textcolor{losscolor}{(-5.5\%)}} & 16.7 {\scriptsize\textcolor{losscolor}{(-46.5\%)}} & 13.3 {\scriptsize\textcolor{losscolor}{(-53.7\%)}} & 2.9 {\scriptsize\textcolor{losscolor}{(-92.8\%)}} & 22.3 {\scriptsize\textcolor{losscolor}{(-40.8\%)}} & 14.0 {\scriptsize\textcolor{losscolor}{(-68.2\%)}} & 21.9 {\scriptsize\textcolor{losscolor}{(-46.8\%)}} \\
TIES+DARE & 50.8 {\scriptsize\textcolor{losscolor}{(-22.6\%)}} & 13.9 {\scriptsize\textcolor{losscolor}{(-55.4\%)}} & 15.2 {\scriptsize\textcolor{losscolor}{(-47.0\%)}} & 2.9 {\scriptsize\textcolor{losscolor}{(-92.8\%)}} & 21.7 {\scriptsize\textcolor{losscolor}{(-42.4\%)}} & 12.0 {\scriptsize\textcolor{losscolor}{(-72.7\%)}} & 19.4 {\scriptsize\textcolor{losscolor}{(-52.9\%)}} \\
WIDEN & 41.7 {\scriptsize\textcolor{losscolor}{(-36.4\%)}} & 11.8 {\scriptsize\textcolor{losscolor}{(-62.2\%)}} & 18.0 {\scriptsize\textcolor{losscolor}{(-37.3\%)}} & 2.3 {\scriptsize\textcolor{losscolor}{(-94.3\%)}} & 17.6 {\scriptsize\textcolor{losscolor}{(-53.3\%)}} & 10.0 {\scriptsize\textcolor{losscolor}{(-77.3\%)}} & 16.9 {\scriptsize\textcolor{losscolor}{(-59.0\%)}} \\
\midrule
NeuronMerge & 60.7 {\scriptsize\textcolor{losscolor}{(-7.5\%)}} & \textbf{35.4} {\scriptsize\textcolor{gaincolor}{(+13.5\%)}} & 18.5 {\scriptsize\textcolor{losscolor}{(-35.5\%)}} & 30.9 {\scriptsize\textcolor{losscolor}{(-22.8\%)}} & \textbf{51.3} {\scriptsize\textcolor{gaincolor}{(+36.1\%)}} & 56.0 {\scriptsize\textcolor{gaincolor}{(+27.3\%)}} & 42.1 {\scriptsize\textcolor{gaincolor}{(+2.2\%)}} \\
AIM & 62.6 {\scriptsize\textcolor{losscolor}{(-4.6\%)}} & 31.2 {\scriptsize\textcolor{gaincolor}{(+0.0\%)}} & 20.9 {\scriptsize\textcolor{losscolor}{(-27.2\%)}} & 18.1 {\scriptsize\textcolor{losscolor}{(-54.8\%)}} & 42.3 {\scriptsize\textcolor{gaincolor}{(+12.2\%)}} & 46.0 {\scriptsize\textcolor{gaincolor}{(+4.5\%)}} & 36.9 {\scriptsize\textcolor{losscolor}{(-10.4\%)}} \\
\rowcolor{highlight}\method(ours) & 64.2 {\scriptsize\textcolor{losscolor}{(-2.1\%)}} & {28.5} {\scriptsize\textcolor{losscolor}{(-8.7\%)}} & \textbf{22.0} {\scriptsize\textcolor{losscolor}{(-23.3\%)}} & \textbf{40.3} {\scriptsize\textcolor{gaincolor}{(+0.8\%)}} & \textbf{51.3} {\scriptsize\textcolor{gaincolor}{(+36.1\%)}} & \textbf{68.0} {\scriptsize\textcolor{gaincolor}{(+54.5\%)}} & \textbf{45.7} {\scriptsize\textcolor{gaincolor}{(+10.9\%)}} \\
\bottomrule
\end{tabular}

\caption{Results with Qwen2.5-7B experts. Metrics follow Table~\ref{tab:qwen3_results}. Parentheses show relative change compared to BEST-of-Three for each aggregate.}
\label{tab:qwen25_results}

\end{table*}

\subsection{Experiment Setup}
\label{sec:exp_setup}

\paragraph{Expert models.}
We adopt Qwen3-8B \citep{yang2025qwen3} as the primary backbone architecture and merge three benchmark-specialized SFT experts released in the \textsc{Simia} framework \citep{li2025simia}, namely Simia-Tau-SFT-Qwen3-8B, Simia-OfficeBench-SFT-Qwen3-8B, and Simia-AgentBench-SFT-Qwen3-8B.
Each expert is fine-tuned on synthesized multi-turn trajectories with benchmark-specific tool interactions, including airline and retail tool calls in a $\tau^2$-Bench-style environment \citep{barres2025tau2bench}, OfficeBench multi-application workflows, and AgentBench-style operating system and WebShop tasks.
For additional comparisons, we also evaluate an expert pool based on Qwen2.5-7B trained with the same \textsc{Simia} recipe.

\paragraph{Baselines.}
We compare \method with strong training-free weight-space merging baselines, including uniform averaging, Model Stock~\citep{jang2024model}, task arithmetic~\citep{ilharco2023task_arithmetic}, TIES~\citep{yadav2023ties}, and TIES+DARE~\citep{yu2024languagemodelssupermario} (all implemented in \textsc{MergeKit}~\citep{goddard2024arcee}), as well as WIDEN~\citep{yu2024widen}, AIM~\citep{aim2025}, and NeuronMerge~\citep{neuronmerge2025}. For detailed baseline settings, see Appendix~\ref{sec:baselines}.

\paragraph{Benchmarks.}
To evaluate the generalization capability of our method, we conduct experiments under both in-domain and out-of-domain settings: 1) \textbf{In-domain benchmarks} include $\tau$-bench~\citep{yao2024taubench}, OfficeBench~\citep{wang2024officebench}, WebShop, and Operating System (Both from AgentBench~\citep{liu2023agentbench})) \textbf{Out-of-domain benchmarks} include DB-bench~\citep{zheng2025lifelongagentbench} and AlfWorld~\citep{shridhar2021alfworld}. Further details on benchmark composition and settings are provided in Appendix~\ref{sec:benchmarks}.

\paragraph{Calibration data and role spans.}
To compute role-conditioned saliency (Section~\ref{sec:selection}), we construct a calibration set that is disjoint from all evaluation and test splits, containing a total of 699 tasks and 1240 trajectories. This calibration set is used solely for forward-pass activation tracing without gradient updates. Details of its composition are provided in Appendix~\ref{sec:calibration}.

\subsection{Main Results}
Tables~\ref{tab:qwen3_results} and~\ref{tab:qwen25_results} summarize the overall results on the Qwen3-8B and Qwen2.5-7B expert pools, from which we draw the following observations:

(1) \textbf{\method yields the strongest single merged generalist across both expert pools.}
It is the only approach that consistently surpasses the BEST-of-Three oracle on both backbones.
We attribute this to \method's two-stage design: AOS-based backbone selection avoids starting from an unstable merge, and the subsequent localized repair targets only the remaining suite-specific deficiencies.

(2) \textbf{Weight-space merging is highly brittle in interactive agent suites.}
Across both backbones, common merge operators show pronounced cross-suite trade-offs, where gains on some environments come with severe regressions on others.
This suggests that global parameter blending can easily perturb role-critical behaviors, and such small deviations may cascade into long-horizon failures in multi-turn trajectories.

(3) \textbf{\method improves cross-environment robustness by isolating role-critical circuits and mitigating negative transfer.}
Compared to both weight-space and activation-aware baselines, \method tends to better preserve performance on role-sensitive suites while retaining strong out-of-domain generalization.
This is mainly due to (i) role-conditioned tracing that focuses saliency on benchmark-critical spans, and (ii) conflict-aware protection during transplantation that prevents overwriting neurons needed by other environments, thereby reducing destructive interference.

\subsection{Ablation Study}
We ablate \method to validate the contribution of each component and to characterize robustness and practical overhead in interactive agent suites.
Our design targets two failure modes highlighted in Section~\ref{sec:intro}: (i) backbone instability across weight-space merge operators, and (ii) destructive interference in multi-turn trajectories, where small errors on role-critical spans (tool calls, action serialization, structured outputs) can cascade into repeated failures.
Unless otherwise stated, AOS and saliency statistics are computed on the disjoint calibration set (Section~\ref{sec:exp}), without using evaluation data.

\paragraph{Effectiveness of AOS as a lightweight proxy for backbone quality.}
\begin{wrapfigure}[20]{r}{0.48\columnwidth}
  \centering
  \vspace{-10pt}
  \includegraphics[width=\linewidth]{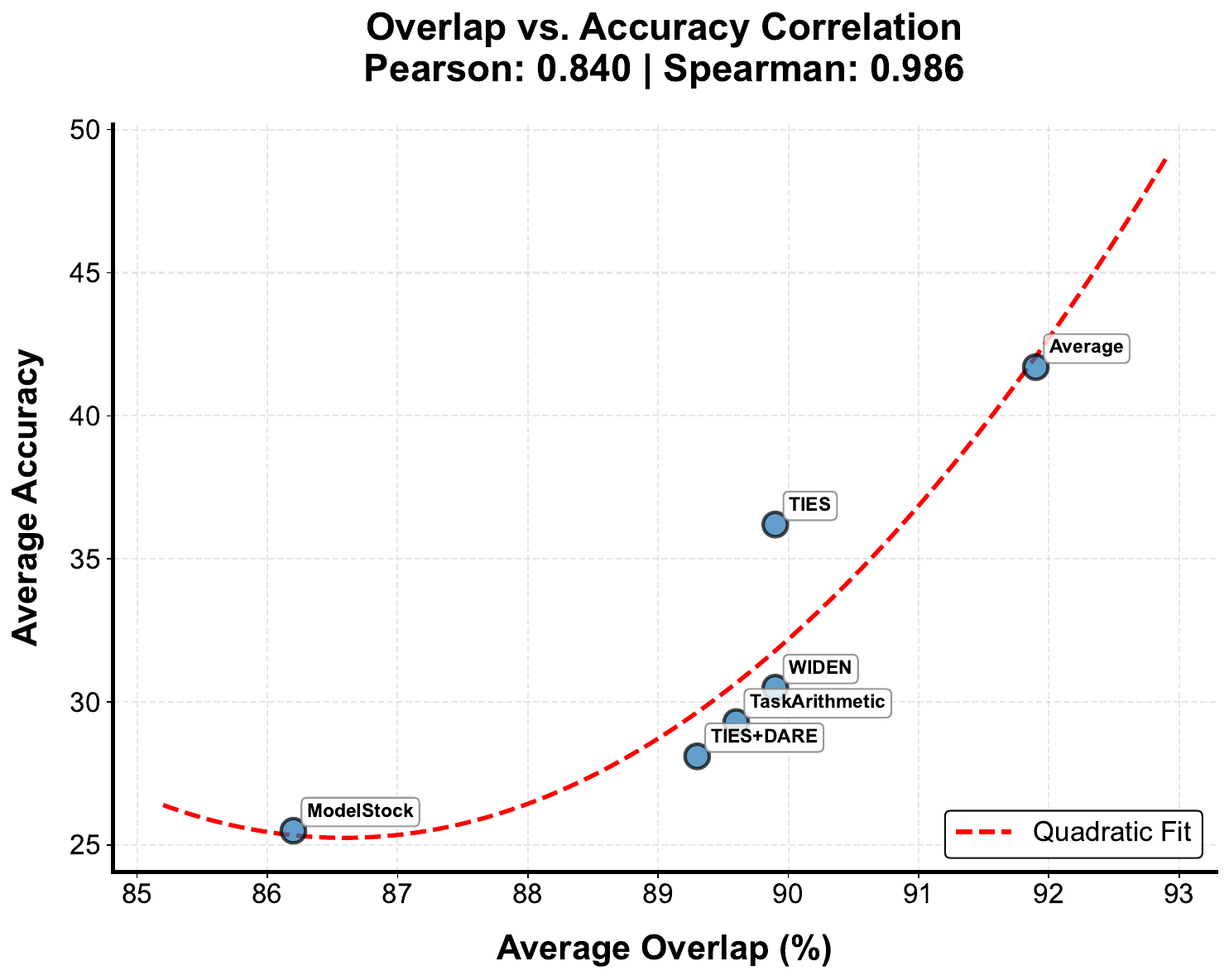}
  \caption{AOS correlates positively with overall performance (\textsc{Avg}) across candidate merge backbones on Qwen3-8B, enabling lightweight initialization selection.}
  \label{fig:aos_correlation}
\end{wrapfigure}
The goal of AOS is to provide a training-free, benchmark-agnostic selection signal that correlates with downstream cross-suite performance, avoiding full interactive evaluation for every merge candidate.
Figure~\ref{fig:aos_correlation} shows a positive relationship between AOS (measured on calibration trajectories) and overall performance (\textsc{Avg}) across candidate backbones.
In our candidate pools, selecting the highest-AOS backbone identifies the best-performing initialization in hindsight: Average for Qwen3-8B and Model Stock for Qwen2.5-7B.
These results support AOS as a practical criterion for reliably choosing a strong starting point before any neuron-level intervention, which is particularly important given the high variability of merge operators in agent environments.

\begin{figure*}[t]
  \centering
  \includegraphics[width=\textwidth]{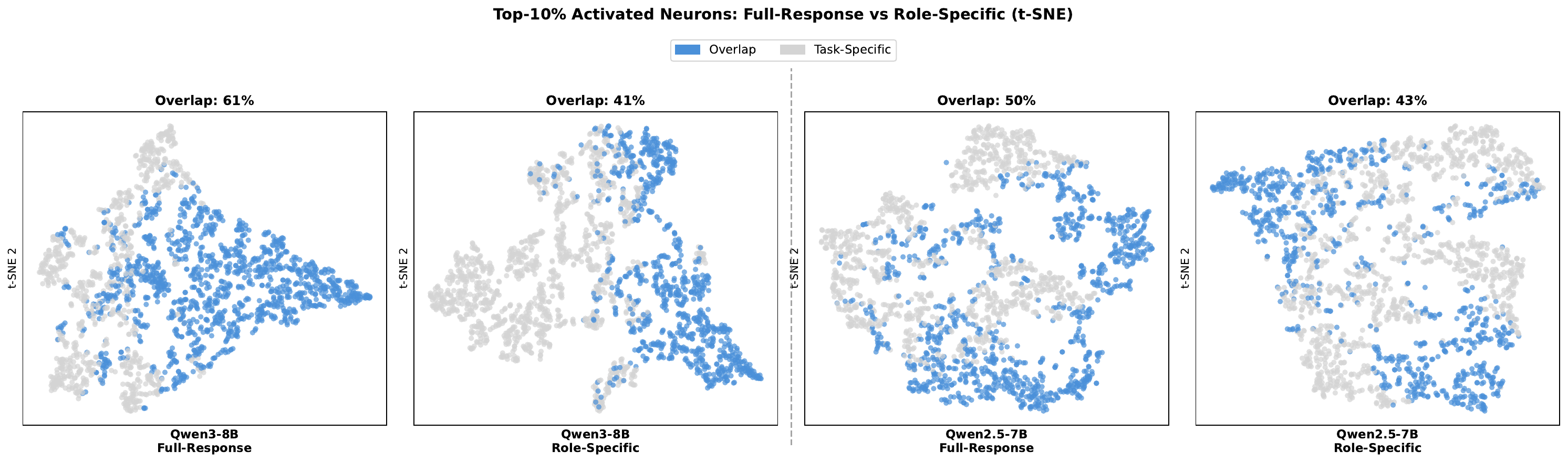}
  \caption{Role-conditioned tracing reduces cross-benchmark overlap of salient neurons. We visualize top-$10\%$ salient neurons and highlight neurons shared across benchmark-specific sets. Compared to full-response tracing, role-conditioned tracing yields lower overlap, indicating reduced cross-environment entanglement of the traced circuits.}
  \label{fig:role_overlap}
\end{figure*}

\paragraph{Role segmentation reduces cross-benchmark interference.}
A core motivation of \method is that agent generalization is often bottlenecked by failures on role-critical spans (e.g., tool-call formats or structured final answers), rather than generic language tokens.
To test whether role-conditioning makes the traced neuron sets more benchmark-specific, we compare role-conditioned tracing against a role-agnostic variant that computes saliency over all response tokens.
Figure~\ref{fig:role_overlap} visualizes the top-$10\%$ salient neurons and highlights neurons shared across benchmark-specific sets.
Role-conditioned tracing yields substantially lower cross-benchmark overlap: the overlap rate drops from $61\%$ to $41\%$ on Qwen3-8B, and from $50\%$ to $43\%$ on Qwen2.5-7B.
This suggests that restricting tracing to benchmark-critical role spans produces more specialized neuron sets, which is desirable for localized transplantation: fewer shared neurons implies fewer accidental edits to capabilities needed by other environments.

\begin{wrapfigure}[16]{r}{0.48\columnwidth}
  \centering
  \vspace{-10pt}
  \includegraphics[width=\linewidth]{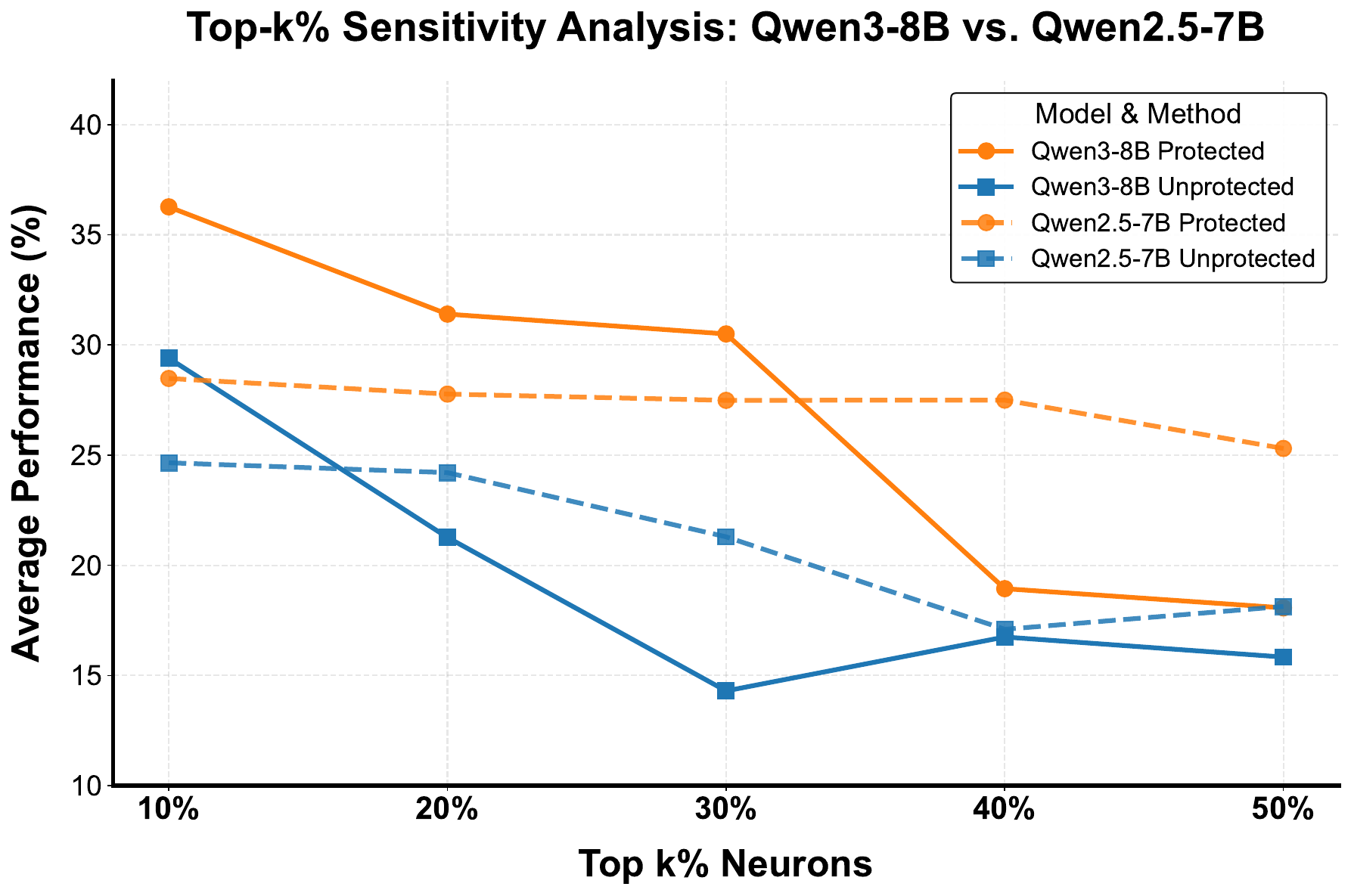}
  \caption{Top-$k$ sensitivity analysis. Conflict-aware protection remains stronger as $k$ increases, indicating improved robustness to the saliency threshold.}
  \label{fig:topk_curve}
\end{wrapfigure}

\paragraph{Conflict-aware protection improves robustness.}
Neuron transplantation can repair benchmark-specific regressions, but directly transplanting all donor-salient neurons may overwrite neurons that are also important for other environments, leading to negative transfer.
\method mitigates this risk via conflict-aware set subtraction (Section~\ref{sec:transplantation}), which removes donor neurons that overlap with the aggregated salient set from the remaining benchmarks.
We ablate this protection by comparing \method against an unprotected variant and sweeping the per-layer top-$k$ fraction used to define role-salient neurons.
Figure~\ref{fig:topk_curve} shows that the unprotected variant is more sensitive to $k$: performance decreases more rapidly as $k$ grows, whereas conflict-aware protection yields consistently higher performance and a flatter degradation trend across a wide range of $k$.
Overall, the results suggest that conflict-aware subtraction improves robustness to the choice of $k$ and helps limit negative transfer when the intervention scope increases.

\paragraph{Generalization metrics beyond \textsc{Avg}.}
To better characterize cross-environment generalization, we report two robustness-oriented summaries in addition to \textsc{Avg}: Worst-suite (\textsc{WS}), the minimum over the six benchmark aggregates, and an oracle-normalized harmonic mean (\textsc{RHM}), which emphasizes balanced performance across suites.
Table~\ref{tab:gen_summaries} shows that the AOS-selected initialization is dominated by its weakest suite, resulting in low \textsc{WS}.
\method improves this worst-case robustness while also increasing \textsc{Avg}, raising \textsc{WS} from 19.1 to 28.5 on Qwen3-8B and from 16.4 to 22.0 on Qwen2.5-7B, and substantially improving \textsc{RHM}.
Overall, these summaries indicate that \method yields a more balanced generalist model that approaches the oracle selector more uniformly by alleviating the weakest-suite bottleneck.

\begin{table}[t]
\centering
\small
\setlength{\tabcolsep}{4.0pt}
\renewcommand{\arraystretch}{1.08}
\begin{tabular}{l l c c c}
\toprule
\textbf{Backbone} & \textbf{Model} & \textsc{Avg} (\%) & \textsc{WS} (\%) & \textsc{RHM} \\
\midrule
Qwen3-8B & AOS-selected  & 41.8 & 19.1 & 0.84 \\
Qwen3-8B & \method & \textbf{44.6} & \textbf{28.5} & \textbf{0.98} \\
\midrule
Qwen2.5-7B & AOS-selected & 43.5 & 16.4 & 0.95 \\
Qwen2.5-7B & \method & \textbf{45.7} & \textbf{22.0} & \textbf{1.04} \\
\bottomrule
\end{tabular}
\caption{Generalization summaries for the AOS-selected initialization backbone and \method. \textsc{Avg} is the unweighted mean over six benchmark aggregates. \textsc{WS} is the minimum over the six aggregates. \textsc{RHM} is the harmonic mean of the six aggregate scores normalized by BEST-of-Three. For Qwen3-8B, the AOS-selected initialization is Average; for Qwen2.5-7B, it is Model Stock.}
\label{tab:gen_summaries}
\end{table}

\paragraph{Failure analysis on role-critical errors.}
A key motivation of ARM is that cross-environment failures in interactive agents are often triggered by localized errors on role-critical spans that can cascade across multi-turn trajectories. To make this failure mode measurable, we leverage the benchmark-specific deterministic parsers already used in our pipeline to identify these spans (tool-call spans for $\tau$-bench, final-answer JSON spans for OfficeBench, and action schema  spans for AgentBench), and we inspect representative episodes where the AOS-selected merged backbone fails due to span-level violations. Compared to the AOS-selected backbone, ARM typically repairs the earliest blocking violation, allowing subsequent tool execution to proceed with minimal changes to the remaining trajectory. We provide side-by-side trajectory excerpts and parser-flagged error annotations in Appendix~\ref{app:case_study}.

\paragraph{Efficiency and overhead.}
\method is training-free and only requires forward-pass activation tracing on a lightweight calibration set; detailed compute, storage, and edit locality statistics are reported in Appendix~\ref{app:efficiency}.

%% file: section/conclusion.tex
\section{Conclusion}
\label{sec:concl}
We presented \textbf{Agent-Role Merging (ARM)}, a training-free framework for consolidating benchmark-specialized LLM agents into a single generalist checkpoint.
ARM addresses two failure modes of agentic model merging: (i) instability across weight-space merge operators, and (ii) destructive interference on role-critical behaviors in multi-turn trajectories.
To this end, ARM selects a strong merged initialization using an Activation-Overlap Score computed from role-conditioned activation tracing, and then performs conflict-aware transplantation of a small set of role-salient MLP neurons to repair weak environments while protecting capabilities needed elsewhere.
Across both Qwen3-8B and Qwen2.5-7B expert pools, ARM achieves the best overall merged model and substantially improves worst-suite robustness.
These results suggest that targeting role-critical circuits enables localized, training-free edits that mitigate negative transfer in interactive agent suites.

%% file: section/discussion.tex
\section{Limitations}
\label{sec:limit}

ARM is training-free, but it makes several assumptions that limit its applicability and leave room for future work.
First, ARM requires access to homologous expert checkpoints that share the same architecture and tokenizer; it does not directly apply to merging heterogeneous model families or black-box APIs.

Second, ARM relies on activation-level signals to identify role-salient circuits, yet diagnostics tailored to multi-turn interactive agent behaviors remain relatively under-explored. 
Future advances in activation-based interpretability for agentic settings would likely enable more accurate interventions and further improve performance.

%% file: section/appendix.tex
\section{Detailed Experiment Settings}
\subsection{Baseline Settings}
\label{sec:baselines}
We use publicly available implementations for all baselines whenever possible. Hyperparameters are set as follows:

\begin{itemize}
\item \textbf{Model Stock}~\citep{jang2024model}: we follow the global coefficient setting with \texttt{filter\_wise=false}, as recommended in the original paper.
\item For all other methods—including uniform averaging, task arithmetic~\citep{ilharco2023task_arithmetic}, TIES~\citep{yadav2023ties}, WIDEN~\citep{yu2024widen}, AIM~\citep{aim2025}, and NeuronMerge~\citep{neuronmerge2025}—we use the default hyperparameters provided by the respective official or paper-reproduced implementations.
\end{itemize}

No benchmark-specific hyperparameter tuning is performed for any baseline.
\subsection{Benchmark Settings}
\label{sec:benchmarks}
We use the official evaluation harness for each suite.
For $\tau$-bench, the user simulator is deterministic with GPT-4.1 at temperature 0, while the evaluated agent uses temperature 0.2 with top-$p{=}1.0$ and fixed seeds to control task-order shuffling and sampling.
For OfficeBench, AgentBench, DB-bench, and AlfWorld, the benchmark defaults are used with temperature 0.7 and top-$p{=}1.0$.
The maximum number of new tokens is set to 512 for OfficeBench and 1024 for AgentBench, DB-bench, and AlfWorld.
For AlfWorld, we use the standard unseen split with a maximum of 35 steps and 1-shot prompting.

We list the evaluation benchmarks used in this work in Table~\ref{tab:benchmark_stats}. For benchmarks with multiple subsets (i.e., $\tau$-bench and OfficeBench), we report the macro-averaged results across all subsets.
\begin{table}[htbp]
    \centering
    \label{tab:dataset_stats}
    \scalebox{0.8}{
    \begin{tabular}{cccc}
        \toprule
        \textbf{Domain} & \textbf{Benchmark} & \textbf{Subset} & \textbf{\# Tasks} \\
        \midrule
        % In-domain section
        \multirow{6}{*}{In-domain} & \multirow{2}{*}{$\tau$-bench} & Airline & 50 \\
                                  &                               & Retail  & 115 \\
        \cmidrule(lr){2-4}
                                  & \multirow{2}{*}{OfficeBench}  & 2-apps   & 51 \\
                                  &                               & 3-apps   & 55 \\
        \cmidrule(lr){2-4}
                                  &  WebShop  & --  & 200 \\
                                  &   Operating System   & --   & 144 \\
        \midrule
        % Out-of-domain section
        \multirow{2}{*}{Out-of-domain} & DB-bench                   & --      & 300 \\
                                     & AlfWorld                     & --      & 50 \\
        \bottomrule
    \end{tabular}
    }
    \caption{Statistics of In-domain and Out-of-domain Evaluation Datasets.}
    \label{tab:benchmark_stats}
\end{table}

\subsection{Calibration Set Settings}
To compute role-conditioned saliency (Section~\ref{sec:selection}), a small calibration set $\calib$ is constructed from splits that are disjoint from our test set. The composition of the calibration set is listed in Table~\ref{tab:calibration}. Deterministic, benchmark-specific parsers are used to trace benchmark-critical spans, including tool-call spans for $\tau$-bench, final-answer JSON spans for OfficeBench, and action schema and argument spans for AgentBench.

\label{sec:calibration}
\begin{table}[htbp]
\centering
\scalebox{0.85}{
\begin{tabular}{lcc}
\toprule
\textbf{Dataset} & \textbf{\Tasks} & \textbf{\Trajectories} \\
\midrule
$\tau$-bench Retail (train)      & 500 & 500 \\
OfficeBench 1-app                & 93  & 372 \\
WebShop (dev)         & 26  & 208 \\
Operating System (dev)              & 80  & 160 \\
\midrule
\textbf{Total}                   & \textbf{699} & \textbf{1240} \\
\bottomrule
\end{tabular}
}
\caption{Composition of the calibration set. To balance the impact of datasets with different sizes, we sample a varying number of trajectories for each dataset, as indicated by the ratio of trajectories to tasks.}
\label{tab:calibration}
\end{table}

\section{Case Studies: Repairing Role-Critical Failure Cascades}
\label{app:case_study}

\paragraph{Setup.}
We analyze representative failure cases to illustrate how merge-induced deviations on role-critical spans can cascade into long-horizon failures in interactive environments.
We focus on suites whose role-critical spans are deterministically identifiable by benchmark-specific parsers used in our pipeline: final-answer JSON spans for OfficeBench, tool-call spans for $\tau$-bench, and action schema spans for OS and WebShop.
Unless otherwise noted, we compare the AOS-selected initialization backbone against ARM under the same decoding and evaluation settings.

\subsection{OfficeBench: Structured Output (JSON) Violations}
\label{app:case_officebench}

\paragraph{Span-level violation rate.}
Table~\ref{tab:officebench_invalid_json} reports the fraction of episodes with invalid final structured outputs that cannot be parsed by the evaluator.
ARM reduces invalid episodes from 8.5\% to 4.7\%.

\begin{table}[t]
\centering
\small
\setlength{\tabcolsep}{5pt}
\renewcommand{\arraystretch}{1.08}
\begin{tabular}{lcc}
\toprule
\textbf{Model} & \textbf{Invalid Episodes} & \textbf{Rate} \\
\midrule
Backbone (Model Stock) & 9/106 & 8.5\% \\
ARM  & 5/106 & 4.7\% \\
\bottomrule
\end{tabular}
\caption{OfficeBench invalid final JSON episode rate (Qwen2.5-7B pool).}
\label{tab:officebench_invalid_json}
\end{table}

\paragraph{Representative failure modes.}
Across failures, the backbone commonly violates the required action/answer structure by nesting a JSON action inside a string field, emitting the entire \texttt{<think><answer>} template as a string, or producing malformed escapes that break JSON parsing, often followed by an invalid got\_stuck action.

\begin{table}[t]
\centering
\small
\setlength{\tabcolsep}{4pt}
\renewcommand{\arraystretch}{1.08}
\begin{tabular}{p{0.20\linewidth} p{0.74\linewidth}}
\toprule
\textbf{Error Type} & \textbf{Description (example)} \\
\midrule
JSON nesting & Action JSON is wrapped as a quoted string inside a field (e.g., \texttt{"action": "\{...\}"}). \\
Template confusion & The model outputs \texttt{<think>...</think><answer>...</answer>} as a literal string field. \\
Escape error & Invalid escape sequences cause JSON parsing failure (e.g., \texttt{Invalid \textbackslash\textbackslash escape}). \\
Crash action & The model emits an invalid fallback action (e.g., \texttt{got\_stuck}) that violates the required schema. \\
\bottomrule
\end{tabular}
\caption{Common structured-output error patterns on OfficeBench.}
\label{tab:officebench_error_patterns}
\end{table}

\paragraph{Case studies.}
We show three representative OfficeBench episodes. In each, the backbone fails due to a structured-output violation, while ARM preserves the required schema and completes the workflow.

\begin{casebox}{OfficeBench Task 2-14: reward meetings for top-3 students}
\textbf{Backbone (Model Stock) -- Fail}\\[-0.25em]
\begin{lstlisting}
{"action": " {\"app\": \"shell\", \"action\": \"command\",
  \"command\": \"sort -nr -k2,2 score.xlsx | head -3\"} "}
{"action": "{\"app\": \"system\", \"action\": \"got_stuck\"}"}
[Error] Malformed action string. Expected <think>...</think><answer>[JSON]</answer>.
Result: Fail
\end{lstlisting}

\tcblower

\textbf{ARM -- Success}\\[-0.25em]
\begin{lstlisting}
{"app":"excel","action":"read_file","file_path":".../score.xlsx"}
{"app":"calendar","action":"create_event","user":"Alice", ...}
{"app":"calendar","action":"create_event","user":"David", ...}
{"app":"calendar","action":"create_event","user":"Carol", ...}
{"app":"system","action":"finish_task","answer":"None"}
Result: Success
\end{lstlisting}
\end{casebox}
\captionof{figure}{OfficeBench Task 2-14. The backbone violates the required structured format and crashes; ARM preserves valid actions and completes the multi-app workflow.}
\label{fig:case_officebench_214}

\begin{casebox}{OfficeBench Task 3-49: pair students and notify via calendar + email}
\textbf{Backbone (Model Stock) -- Fail}\\[-0.25em]
\begin{lstlisting}
{"action":"<think>...</think><answer> {\"app\":\"system\",
\"action\":\"switch_app\",\"target_app\":\"excel\"} </answer>"}
[Error] Malformed action string. Expected <think>...</think><answer>[JSON]</answer>.
Result: Fail
\end{lstlisting}

\tcblower

\textbf{ARM  -- Success}\\[-0.25em]
\begin{lstlisting}
{"app":"excel","action":"read_file","file_path":".../schedule.xlsx"}
{"app":"calendar","action":"create_event","user":"Alice", ...}
{"app":"calendar","action":"create_event","user":"Carol", ...}
{"app":"email","action":"send_email","sender":"Alice", ...}
{"app":"email","action":"send_email","sender":"Carol", ...}
{"app":"system","action":"finish_task","answer":"None"}
Result: Success
\end{lstlisting}
\end{casebox}
\captionof{figure}{OfficeBench Task 3-49. The backbone emits the full think-answer template as a literal string; ARM produces valid structured actions and finishes the task.}
\label{fig:case_officebench_349}

\begin{casebox}{OfficeBench Task 3-7: invalid escape sequence}
\textbf{Backbone (Model Stock) -- Fail}\\[-0.25em]
\begin{lstlisting}
[Error] Invalid \\escape ... Malformed action!
{"action":"{\"app\":\"system\",\"action\":\"got_stuck\"}"}
Result: Fail
\end{lstlisting}

\tcblower

\textbf{ARM -- Success}\\[-0.25em]
\begin{lstlisting}
{"app":"excel","action":"read_file","file_path":".../students.xlsx"}
{"app":"word","action":"write_to_file","file_path":".../admission.docx", ...}
{"app":"email","action":"send_email","recipient":"jennifer.gonzalez@example.com", ...}
...
Result: Success
\end{lstlisting}
\end{casebox}
\captionof{figure}{OfficeBench Task 3-7. The backbone triggers a JSON parsing error (invalid escape); ARM maintains valid structured outputs and successfully completes multi-recipient execution.}
\label{fig:case_officebench_37}

\subsection{$\tau$-bench: Tool-Call Failure Cascades}
\label{app:case_taubench}

\paragraph{Representative tool-call cascades.}
We present three cases where the backbone either repeats a failing tool call without correcting the underlying issue, makes redundant queries and acts on the wrong target, or omits a required critical tool action.

\begin{casebox}{$\tau$-bench Task 13: error loop after a tool failure}
\textbf{Backbone -- Fail (loop)}\\[-0.25em]
\begin{lstlisting}
update_reservation_passengers -> Error (passengers mismatch)
cancel_reservation -> OK
book_reservation -> Error (payment amount does not add up)
book_reservation -> Error (repeated many times)
Reward: 0.0
\end{lstlisting}

\tcblower

\textbf{ARM -- Success}\\[-0.25em]
\begin{lstlisting}
get_reservation_details -> OK
search_direct_flight -> OK
...
Reward: 1.0
\end{lstlisting}
\end{casebox}
\captionof{figure}{$\tau$-bench Task 13. The backbone enters a repeated error loop after a tool failure; ARM resolves the issue and completes without looping.}
\label{fig:case_taubench_13}

\begin{casebox}{$\tau$-bench Task 31: redundant queries and wrong target}
\textbf{Backbone -- Fail}\\[-0.25em]
\begin{lstlisting}
get_reservation_details called repeatedly on unrelated reservations
cancel_reservation -> cancelled an incorrect booking
Reward: 0.0
\end{lstlisting}

\tcblower

\textbf{ARM -- Success}\\[-0.25em]
\begin{lstlisting}
get_user_details -> OK
cancel_reservation(correct id) -> OK
Reward: 1.0
\end{lstlisting}
\end{casebox}
\captionof{figure}{$\tau$-bench Task 31. The backbone makes redundant queries and cancels the wrong booking; ARM cancels the intended reservation directly.}
\label{fig:case_taubench_31}

\begin{casebox}{OS: command choice affects execution correctness}
\textbf{Backbone (Average) -- Failed}\\[-0.25em]
\begin{lstlisting}
touch logfile.txt
echo "abc" >> logfile.txt   # includes newline
tr -cd '[:alnum:]' < logfile.txt | sort -u | wc -l
Answer: 1 (Wrong)
\end{lstlisting}

\tcblower

\textbf{ARM -- Success}\\[-0.25em]
\begin{lstlisting}
echo -n "abc" > logfile.txt   # no newline
fold -w1 | sort | uniq | wc -l
Answer: 3 (Correct)
\end{lstlisting}
\end{casebox}
\captionof{figure}{OS case study. The backbone fails due to incorrect file handling, while ARM succeeds.}
\label{fig:case_os_example}

\begin{casebox}{$\tau$-bench Task 46: missing a required critical action}
\textbf{Backbone -- Fail}\\[-0.25em]
\begin{lstlisting}
get_user_details -> OK
get_reservation_details -> OK
(terminated without send_certificate)
Reward: 0.0
\end{lstlisting}

\tcblower

\textbf{ARM -- Success}\\[-0.25em]
\begin{lstlisting}
get_user_details -> OK
get_reservation_details -> OK
send_certificate(amount=50) executed
Reward: 1.0
\end{lstlisting}
\end{casebox}
\captionof{figure}{$\tau$-bench Task 46. The backbone omits a required tool action; ARM executes the critical step and completes the task.}
\label{fig:case_taubench_46}

\subsection{OS and WebShop: Action Schema and Execution Errors}
\label{app:case_agentbench}

\paragraph{Validation signals and task outcomes.}
Tables~\ref{tab:validation_metrics} report validation signals (invalid action and task-limit timeouts)  on OS and WebShop.

\begin{table}[H]
\centering
\small
\setlength{\tabcolsep}{4pt}
\begin{tabular}{llcccc}
\toprule
\textbf{Model} & \textbf{Benchmark} & \multicolumn{2}{c}{\textbf{Invalid Action}} & \multicolumn{2}{c}{\textbf{Task Limit}} \\
\cmidrule(lr){3-4} \cmidrule(lr){5-6}
 & & Backbone & ARM & Backbone & ARM \\
\midrule
\multirow{2}{*}{Qwen3-8B} & OS & 2.08 & 1.39 & 12.50 & 8.33 \\
 & WebShop & 0.00 & 0.00 & 0.50 & 3.00 \\
\midrule
\multirow{2}{*}{Qwen2.5-7B} & OS & 0.69 & 0.00 & 15.97 & 18.75 \\
 & WebShop & 0.00 & 0.00 & 5.00 & 0.50 \\
\bottomrule
\end{tabular}
\caption{Validation metrics on AgentBench. ``Invalid Action'' indicates malformed agent outputs rejected by the environment. ``Task Limit'' indicates failure to complete within the maximum step limit. All values are percentages (\%).}
\label{tab:validation_metrics}
\end{table}

\paragraph{Qualitative example.}
Figure~\ref{fig:case_os_example} shows a representative OS episode in which both models emit valid actions, but the backbone executes an imprecise command and returns an incorrect answer.

\paragraph{Summary.}
Across suites, these cases show that many merge failures originate from localized violations on role-critical spans or early tool/action mistakes that derail multi-turn trajectories.
ARM frequently prevents such cascades by preserving required structured formats and executing key tool/action steps more reliably.

\section{Efficiency and Overhead Details}
\label{app:efficiency}
\method is training-free and operates via forward-pass tracing on a lightweight calibration set.
In our setup, AOS-based backbone selection traces activations for six candidate backbones across four in-domain benchmarks, costing $\sim$0.5 GPU-hour per backbone--benchmark pair on a single H20 (about 12 GPU-hours total); once the backbone is selected, the merge and neuron transplantation completes in under 20 minutes.
At the benchmark level, each transplant set is typically small (roughly 2--3\% for $\tau$-bench, OfficeBench, and WebShop).
Activation statistics are stored in compressed NPZ files, requiring less than 500MB total.
Overall, these results indicate that \method can produce a more robust generalist agent with modest one-time calibration cost and targeted neuron-level edits, without any additional training.